\documentclass{article}

\usepackage{arxiv}

\usepackage[utf8]{inputenc} % allow utf-8 input
\usepackage[T1]{fontenc}    % use 8-bit T1 fonts
\usepackage{hyperref}       % hyperlinks
\usepackage{url}            % simple URL typesetting
\usepackage{booktabs}       % professional-quality tables
\usepackage{amsfonts}       % blackboard math symbols
\usepackage{nicefrac}       % compact symbols for 1/2, etc.
\usepackage{microtype}      % microtypography
\usepackage{cleveref}       % smart cross-referencing
\usepackage{lipsum}         % Can be removed after putting your text content
\usepackage{graphicx}
\usepackage[numbers]{natbib}
\usepackage{doi}

\usepackage{adjustbox}
\usepackage{subcaption}
\usepackage{makecell}
\usepackage{stfloats}

\title{Generative AI Use Cases In Real Estate Marketing: Adoption and Constraints in Germany}

\date{}

\newif\ifuniqueAffiliation
\ifuniqueAffiliation % Standard variant of author block
\else
\usepackage{authblk}

\author[1,2]{Victor Kolominsky-Rabas\thanks{\texttt{victor.kolominsky-rabas@fit.fraunhofer.de}}}
\author[1,2]{Leopold Müller}
\author[1]{Felicia Perpina}
\author[1,2]{Niklas Kühl}

\affil[1]{University of Bayreuth, Universitätsstraße 30, 95447 Bayreuth, Germany}
\affil[2]{Fraunhofer FIT, Wittelsbacherring 10, 95444 Bayreuth, Germany}
\fi

\renewcommand{\headeright}{PREPRINT}
\renewcommand{\undertitle}{PREPRINT}
\renewcommand{\shorttitle}{Generative AI Use Cases In Real Estate Marketing}

\hypersetup{
pdftitle={Generative AI Use Cases In Real Estate Marketing: Adoption and Constraints in Germany},
pdfauthor={Victor Kolominsky-Rabas},
}

\begin{document}
\maketitle

\begin{abstract}
Generative artificial intelligence (GenAI) is changing how work is organized and performed. Real estate marketing is a prime example of this, yet evidence of GenAI in real estate agents' day-to-day practice remains scarce. In this work, we report on our insights from a German-based empirical study with eleven semi-structured interviews. GenAI is already utilized across different activities, with marketing communication being the most prominent. Concrete use cases are emergent and unevenly adopted, with writing expos\'e texts being the only widely established one. Interaction is predominantly human-in-the-loop: GenAI drafts, structures, and retrieves, while real estate agents curate, verify, and decide. Constraints stem less from model capability than from integration with listings and documents, data availability, and compliance in sensitive tasks. The study contributes a grounded map of existing and potential use cases and identifies tentative practical implications for adoption.
\end{abstract}

%\include{1_JBR/sections/0-abstract}

% keywords can be removed

\section{Introduction}

Generative artificial intelligence (GenAI) has become widely accessible in everyday and professional work~\citep{von2025identifying}. Reported benefits include creativity support, fewer errors, and productivity gains across sectors~\citep{filz_generative_2024,ghaffari2024generative,mittal2024comprehensive}. Real estate (RE) is information-intensive, yet many processes still depend on scattered listings, heterogeneous property documents, local market knowledge, and manual coordination across portals, agencies, public records, and professional service providers. Prior work describes RE as constrained by weak standardization and fragmented information practices~\citep{bodenbender2019broad,al-haimi_digital_2025,gardes_ally_2025,kriegbaum_chatbots_2024}. This is especially relevant for RE marketing (REM), where agents transform incomplete property information into expos\'es, listings, social media content, client communication, and sales material while maintaining factual accuracy and regulatory compliance. Because REM is shaped by country-specific transaction routines, data infrastructures, and regulation, this study has a single-country focus. As German RE agents (REAs) rely on formal documents such as land register excerpts, cadastral maps, development plans, energy certificates, and notarial transaction processes~\citep{helfrich_erfolgsstrategien_2021}, Germany is an interesting setting. They also operate under European data-protection and AI-governance requirements. At the same time, industry estimates suggest substantial GenAI potential in German RE, while only a small share of firms report high digital maturity~\citep{fitzpatrick_power_2023,zentraler_immobilien_ausschuss_e_v_zia_ey_2024}. Thus, the field is characterized by uneven experimentation and limited routine organizational integration.

Evidence on how REAs actually use GenAI remains scarce. \citet{kriegbaum_chatbots_2024} provide an U.S.-based account of REA sentiment, but RE practices vary across countries and cannot simply be transferred to Germany. We complement this work with a task-level analysis that distinguishes current from potential REM use cases and links them to adoption constraints. We ask:

\begin{itemize}
    \item[RQ1] Which use cases of GenAI are already being applied by real estate agents in Germany?
    \item[RQ2] Which potential use cases does GenAI offer for real estate marketing?    
\end{itemize}

We conduct eleven semi-structured interviews with REAs from different agencies and analyze them inductively following~\citet{gioia_seeking_2013}. The study contributes, first, a grounded map of current and potential GenAI use cases in German REM. Second, it shows that interaction is predominantly human-in-the-loop: GenAI drafts, structures, and retrieves information, while REAs verify outputs, make judgments, and remain accountable. Third, it identifies adoption challenges around listings, regulatory documents, data availability, and compliance-sensitive tasks. In doing so, it extends U.S. evidence by showing which patterns travel across contexts, such as the centrality of marketing content, and which become more salient in fragmented and strongly regulated settings, such as cross-portal aggregation, regulatory document understanding, and compliance-aware screening.
\section{Background and related work}

This section introduces established tasks of REAs, which serve as the practical framework for situating GenAI use cases in REM. It then defines GenAI and finally presents related work on the application of AI and GenAI in REM.

\subsection{Core activities of real estate agents}

REA activities follow the transaction flow~\citep{helfrich_erfolgsstrategien_2021}: \textit{acquisition} (contacting owners intending to sell via channels such as flyers, events, or partnerships), \textit{intake} (capturing key property data, taking professional photos, assembling documentation like land register excerpts and cadastral maps), \textit{property analysis} (market and location analyses based on intake data), \textit{marketing} (expos\'es, social media, listings, flyers; potentially home staging and outreach), viewings (pre-selection and tours), \textit{contract closing} (reviewing notarial drafts and accompanying clients), and \textit{handover}, as depicted in \Cref{fig:mapping}.

\subsection{Generative artificial intelligence in real estate marketing}

AI refers to systems that interpret data, learn from it, and adapt their behavior to achieve specific goals~\citep{haenlein2019brief}. GenAI is a more specific class of AI that generates or transforms content such as text, images, audio, or video based on learned patterns~\citep{feuerriegel_generative_2024}. In this study, a use case is considered GenAI-related when GenAI generates, rewrites, summarizes, interprets, or visually transforms real estate marketing content, or when it orchestrates such steps in an agentic workflow. This boundary is important because several REM tasks also involve established technologies such as data analytics, workflow automation, or rule-based software. We therefore do not assume that every technical step is uniquely GenAI-based. Rather, we focus on tasks that REAs currently perform, or plausibly expect to perform, with GenAI as a substantive component.
Prior RE research still mainly addresses AI-based prediction, optimization, valuation, visualization, or single technical systems rather than GenAI use cases in REA work~\citep{chen_predicting_2024,cheung_real_2024,dombrowski_optimizing_2025,gloria_real-gpt_2025,haurum_real_2024,kucklick2021quantifying,kvet_real_2025,santos2024real,zhao2023real,zhao2024generative,geerts2025performance}. Conceptual work proposes GenAI applications such as listings, virtual staging, routine client communication, screening, lease management, and document generation, but often without empirical validation~\citep{ma2023chatgpt,szumilo2024real}. Other studies discuss skills, education, transparency, discrimination, governance, and market-level effects~\citep{cheung_real_2024,dyason_real_2024,seagraves2024revolutionizing,liu2024racial,wang_impact_2024}. Empirical evidence on GenAI in day-to-day REM remains limited. Existing work shows GenAI-supported social media marketing~\citep{li2025ai}, agentic property-description generation~\citep{wu_ai_2025}, and U.S. practitioner sentiment toward GenAI in real estate~\citep{kriegbaum_chatbots_2024}. \citet{kriegbaum_chatbots_2024} show that marketing material creation is a key adoption catalyst and that professionals associate GenAI with efficiency gains, human intervention, ethical concerns, generational differences, relational work, and future technological development. 
Building on this sentiment-oriented account, we examine how GenAI is incorporated into concrete REM tasks, which applications are already routine rather than anticipated, and how adoption unfolds in the German institutional setting. This shifts the focus from broad adoption sentiment to task-level workplace practice and clarifies which patterns appear robust across contexts and which depend on fragmented and regulated national market structures.
\section{Methodology}

To identify GenAI use cases in REM, we conducted eleven semi-structured interviews \citep{lewis-beck_semistructured_2004} between June and August 2025 (average duration: $29.6$ minutes). Participants were recruited through convenience and snowball sampling~\citep{babbie2020practice,biernacki1981snowball} from different companies across Germany and cover mixed RE activities. Inclusion required active professional use of AI because RQ1 asks for concrete current use cases; non-users are therefore less able to report practiced applications, although their perspectives are important for studying resistance. The sample includes six men and five women with an average of $11.2$ years in the industry and 5--50 closed transactions annually. \Cref{tab:expert-list} summarizes variation in role, employment status, seniority, transaction volume, and self-assessed IT and GenAI skills. Interviews were recorded with consent, transcribed, and manually revised. Participants first described their role and understanding of GenAI before discussing current and potential GenAI use cases. We analyzed the transcripts using an iterative Gioia-style coding procedure \citep{gioia_seeking_2013}. Only passages referring to current or potential GenAI use cases in REM were coded. We derived first-order concepts through open coding, aggregated them into second-order themes, and distilled aggregate dimensions through author workshops. Coding proceeded in two cycles, one for current and one for potential use cases. Because new concepts emerged in the final interviews, full theoretical saturation is not claimed.

\begin{table*}[b]
\centering
\caption{Expert List (EE = Employed, SE = Self-Employed, and confidence scores 0--3).}
\label{tab:expert-list}
\begin{tabular}{cccccccc}
\hline
\textbf{Expert} & \textbf{Agent role} & \textbf{Employment} & \makecell{\textbf{Years in}\\\textbf{industry}} & \textbf{\# agencies} & \textbf{\# transactions} & \makecell{\textbf{IT-}\\\textbf{skills}} & \makecell{\textbf{GenAI-}\\\textbf{skills}} \\
\hline
$\alpha$ & seller & SE & 15 & 1 & 25--30 & 3 & 2--3 \\
$\beta$ & seller & EE & 6 & 1 & 40--50 & 3 & 3 \\
$\gamma$ & seller/buyer & EE & 12 & 2 & 40--50 & 3 & 3 \\
$\delta$ & seller/buyer & SE & 13 & 2 & 30 & 3 & 3 \\
$\varepsilon$ & seller/buyer & EE & 20 & 2 & 50 & 3 & 0 \\
$\zeta$ & trainee & EE & $<1$ & 1 & 50 & 3 & 2 \\
$\eta$ & seller & EE & 10 & 2 & 20--25 & 3 & 2 \\
$\theta$ & seller/buyer & SE & 13 & 3 & 50 & 2 & 2 \\
$\iota$ & seller & SE & 15 & 2 & 10--15 & 3 & 2--3 \\
$\kappa$ & seller & SE & 3 & 2 & 50 & 3 & 3 \\
$\lambda$ & seller & SE & 15 & 1 & 6 & 3 & 2 \\
\hline
\end{tabular}
\end{table*}
\section{Results}

This section reports the empirical findings. We distinguish current use cases, which describe how REAs employ GenAI today, from potential use cases, which REAs consider feasible in the near future. All figures follow the same structure: first-order concepts, second-order themes, and aggregate dimensions from left to right. Upright text denotes current use cases, \textit{italics} denote potential use cases, and * marks potential use cases that some REAs also cited as current.

\subsection{Marketing communication}

\begin{figure*}[thb]
    \centering
	\includegraphics[width=\linewidth]{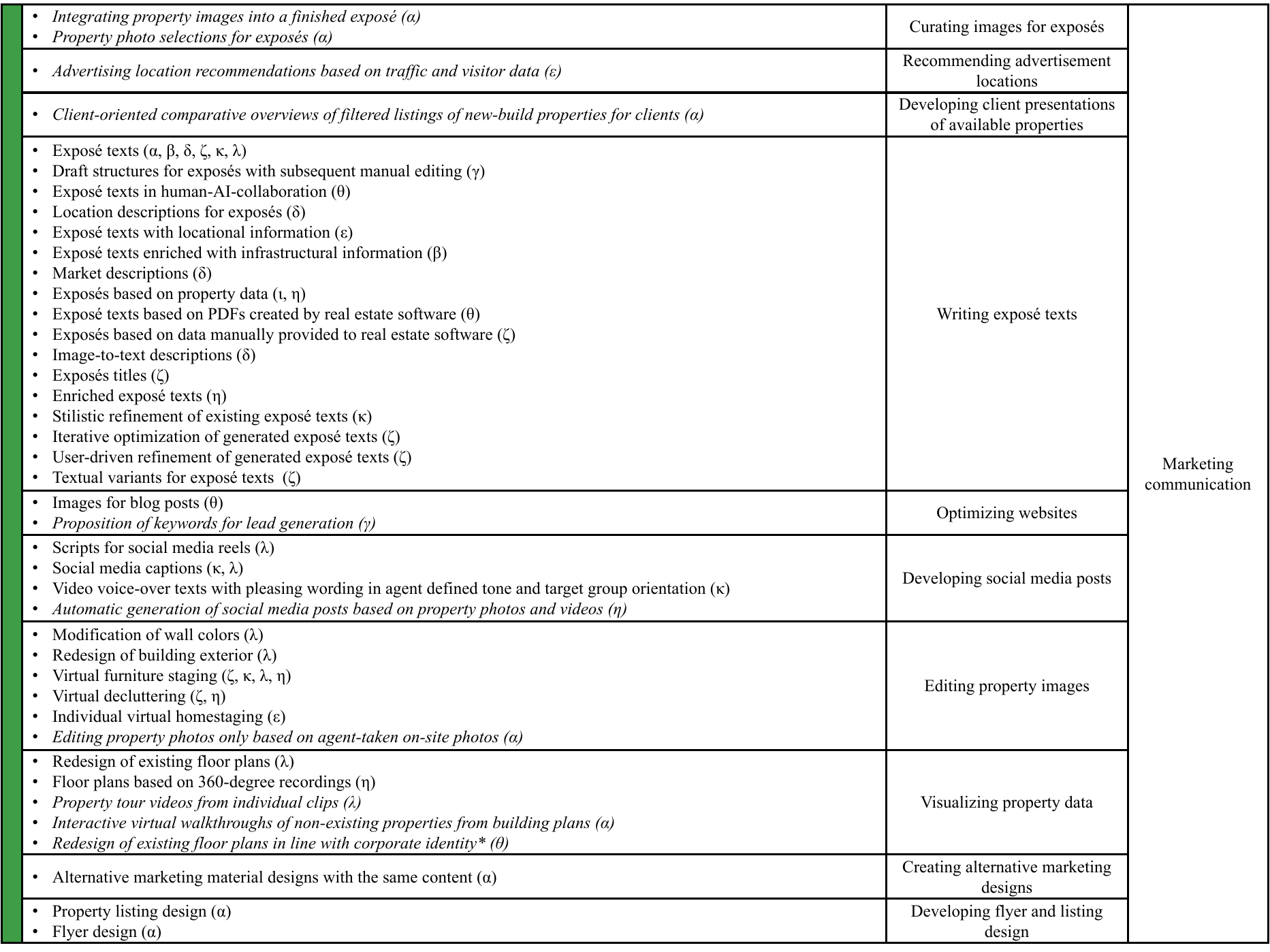}
	\caption{GenAI use cases for marketing communication as derived from interview codes.}
	\label{fig:marketing-communication}
\end{figure*}

As shown in \Cref{fig:marketing-communication}, REAs use GenAI widely for property-facing marketing artifacts such as expos\'es, listings, visuals, websites, and advertisements.

\textbf{Writing expos\'e texts} refers to using GenAI to generate and refine property descriptions. Four REAs use property data to generate complete expos\'e drafts from inputs ranging from textual facts to pre-generated PDFs from internal systems. They also use GenAI for location details, expos\'e titles, stylistic refinement, and structural proposals. Several REAs emphasize collaboration with the model through iterative prompting and human revision. For this theme, all mentions describe current use.
\textbf{Editing property images} involves modifying or enhancing images to improve presentation quality. Current uses include changing wall colors, redesigning exterior elements, virtual staging, decluttering, and tailoring interior styles to client preferences. One REA describes editing photos directly from on-site captures as a potential workflow improvement. Such uses require caution because image edits can misrepresent the property's actual condition and create reputational risks if clients perceive expos\'e images as misleading during viewings.
\textbf{Developing social media posts} includes creating platform-ready marketing content. REAs report current use for writing scripts for reels, captions, and voiceover texts. One REA describes GenAI-generated posts from on-site photos and videos as a potential use case. 
\textbf{Visualizing property data} relates to transforming technical information into accessible visual formats. Current uses include redesigning floor plans and generating floor plans from 360-degree recordings. Potential uses include interactive walk-throughs of unbuilt properties from building plans and virtual tours compiled from short on-site videos. One REA labels floor-plan redesign as potential, while another already practices it, indicating uneven adoption. 
\textbf{Optimizing websites} involves applying GenAI to improve agency web presence. A current example is generating images for blog posts. As a potential use case, REAs suggest GenAI-based keyword proposals to enhance search engine optimization and lead generation. 
\textbf{Curating images for expos\'es} refers to the automated selection and integration of property photos into expos\'es. REAs describe a potential use case in which GenAI chooses suitable images and inserts edited versions directly into the expos\'e, removing manual steps after photography.
\textbf{Developing client presentations of available properties} denotes a potential use case for GenAI-produced presentation material, for example by summarizing and visualizing current listings for clients.
\textbf{Recommending advertisement locations} captures a potential use case for GenAI in identifying where to place advertisements. One example is proposing locations for advertisement walls based on AI-provided information about ``\textit{how many cars pass by there every day, how busy this supermarket [is]}'' (Agent $\epsilon$).
\textbf{Creating alternative marketing designs with the same content} captures the use of GenAI to produce stylistic or layout variations for marketing material while keeping the content unchanged. 
\textbf{Developing flyer and listing designs} covers GenAI assistance in designing flyers and listings. One REA reports current use in which GenAI produces a layout template that streamlines the design process.

\subsection{Acquisition preparation}

\begin{figure*}[thb]
    \centering
	\includegraphics[width=\linewidth]{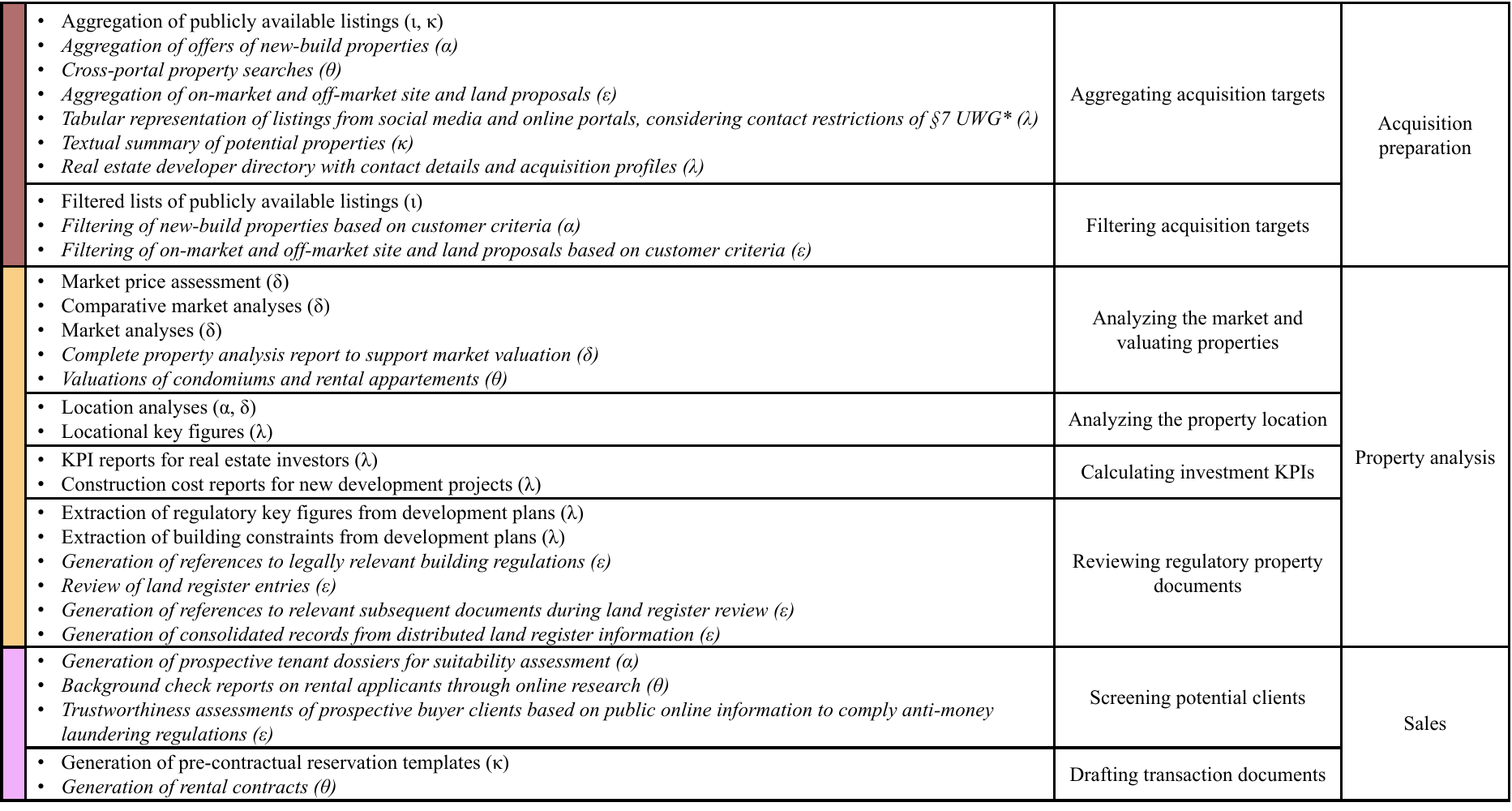}
	\caption{GenAI use cases for acquisition preparation, property analysis and sales as derived from interview codes.}
	\label{fig:acquisition-property-sales}
\end{figure*}

GenAI can support REAs in acquisition preparation (see \Cref{fig:acquisition-property-sales}). Here, acquisition refers to proactively identifying and approaching potential sellers, buyers, landlords, or tenants.

\textbf{Aggregating acquisition targets} refers to using GenAI to compile properties and contacts relevant for upcoming acquisition activities. Two REAs report current aggregation of publicly available listings. Potential uses include compiling new-build properties from the web, searching across portals, matching results to client requirements, producing tabular overviews of listings from social media and online portals, flagging entries that may not be contacted under local regulations, and summarizing candidate properties. Beyond properties, one REA envisions a regional directory of RE developers with contact details and acquisition profiles to reduce preparation time. 
\textbf{Filtering acquisition targets} encompasses GenAI support for narrowing the aggregated pool based on specified criteria, such as client preferences, to create a focused shortlist for outreach.

\subsection{Property analysis}

To market and sell a property, REAs conduct a series of analyses. GenAI supports several of these steps (see \Cref{fig:acquisition-property-sales}).

\textbf{Analyzing the property location} includes cases where GenAI provides information on locational characteristics, such as demographic indicators and local sales activity. 
\textbf{Calculating investment KPIs} concerns quantitative indicators for investment decisions. One REA reports using KPI reports for investors, including gross rental yield and return on equity, which reduces manual calculation effort. REAs also use GenAI to produce construction cost reports for new development projects. Because general-purpose GenAI systems may produce erroneous calculations, KPI use should be understood as drafting or explanation support rather than as a substitute for validated data analytics or spreadsheet-based calculation tools.
\textbf{Reviewing regulatory property documents} refers to support in understanding and analyzing legal and regulatory texts. Current use focuses on development plans, where GenAI extracts regulatory key figures and identifies building constraints. Potential use cases include land register reviews that reference applicable building regulations and subsequent documents, reducing manual searches for REAs. Participants also point to consolidating distributed land register-related documents into a single, navigable view.

\subsection{Sales}

REAs report several sales-related use cases, as depicted in~\Cref{fig:acquisition-property-sales}.

\textbf{Screening potential clients} encompasses GenAI support for evaluating prospective tenants or buyers. A few REAs report current use for scanning publicly available online information to flag risk indicators in support of anti-money laundering obligations. Potential uses include background checks, verification of employment information, assessment of publicly shared content, and applicant dossiers with suitability assessments based on information such as credit scores. While participants expect reduced manual effort and more standardized evaluation, they emphasize that screening involves sensitive personal data and express uncertainty about compliance with the EU AI Act and the General Data Protection Regulation (GDPR). As a result, most screening-related use cases are discussed as potential rather than established practice. 
\textbf{Drafting transactional documents} refers to using GenAI to create documents required in RE transactions. One REA reports drafting non-binding reservation templates as a current use case. Others view the generation of more formal documents, such as rental contracts, as a potential next step. Legal enforceability and liability concerns were noted as reasons to keep humans in the loop even when drafting is automated~\citep{kandipati_systematic_2025,szumilo2024real}.

\subsection{General communication}

\begin{figure*}[thb]
    \centering
	\includegraphics[width=\linewidth]{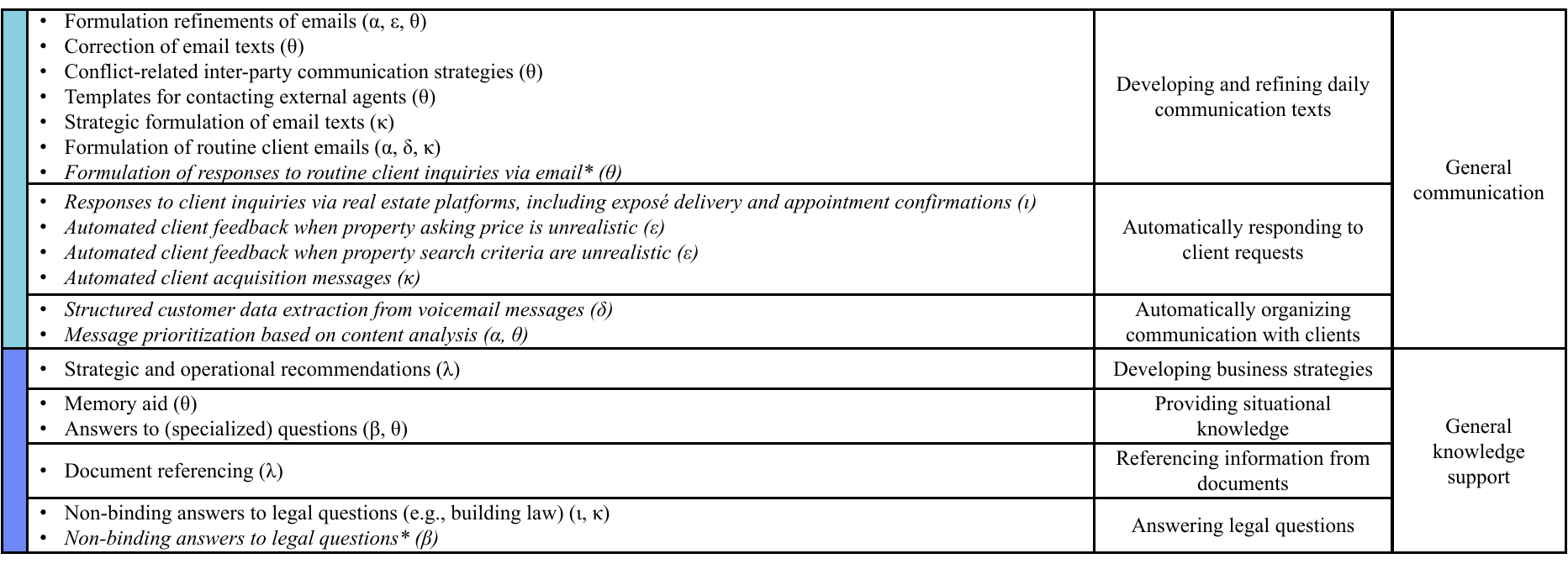}
	\caption{GenAI use cases for general communication and general knowledge support as derived from interview codes.}
	\label{fig:general-tasks}
\end{figure*}

The data also reveal use cases in which GenAI supports REAs in daily work. As shown in \Cref{fig:general-tasks}, these use cases cluster into two domains. In contrast to marketing communication, general communication is not aimed at producing property-facing promotional artifacts but at coordinating client relationships and transaction processes, such as inquiries, appointments, viewings, negotiations, and follow-up.

\textbf{Developing and refining daily communication texts} covers the formulation and strategic enhancement of written and spoken messages. REAs report rephrasing emails, correcting errors, drafting replies to routine client inquiries, and creating conversation templates for arranging viewings or requesting listing details. Participants also use GenAI for strategic guidance in difficult interactions, for example in negotiations or tactical phrasing when interest in a property is limited. 
\textbf{Automatically responding to client requests} encompasses GenAI-generated replies to client requests. One REA expects GenAI to process platform inquiries by delivering expos\'es or confirming appointments. Participants also anticipate automated feedback that flags unrealistic expectations, such as informing owners when asking prices are too high or when criteria cannot be met. One REA suggests that initial acquisition outreach could be automated after a promising property is identified.
\textbf{Automatically organizing communication with clients} refers to using GenAI to structure and prioritize incoming messages. Participants describe potential uses in which GenAI extracts customer data from voicemail, converts unstructured speech into records for documentation and follow-up, and highlights urgent requests.

\subsection{General knowledge support}

The category of general knowledge support includes use cases that are not tied to a single REA task but provide overarching support across different activities (\Cref{fig:general-tasks}). Their relevance for REM lies in the RE-specific content being retrieved or interpreted, such as property documents, building law, development plans, and market or presentation decisions.

\textbf{Developing business strategies} covers requests for strategic and operational recommendations. One REA reports current use by asking GenAI for advice on business questions, including how to present a property more effectively. To tailor outputs, the REA assigns GenAI a specific role, such as a RE assistant with marketing expertise. 
\textbf{Providing situational knowledge} refers to ad hoc information seeking. REAs describe current use as a memory aid for forgotten details and as support for answering general and specific questions. 
\textbf{Referencing information from documents} relates to retrieving and localizing details in lengthy or complex materials. One REA reports current use in which GenAI extracts specific points from documents, such as development plans, and indicates where the relevant content appears, which reduces manual review time. 
\textbf{Answering legal questions} captures preliminary, non-binding support on legal matters. REAs report using GenAI to obtain first answers in areas such as building law, emphasizing reduced effort in navigating legal texts: ``\textit{Before I start leafing through legal codes or researching until I finally have the relevant paragraphs together, using AI is definitely a major advantage.}'' (Agent $\iota$). Another REA mentions this as a potential use case, indicating mixed adoption for this theme~\citep{szumilo2024real}.

\subsection{Code distribution}

\begin{figure*}[tb]
    \centering
	\includegraphics[width=0.7\linewidth]{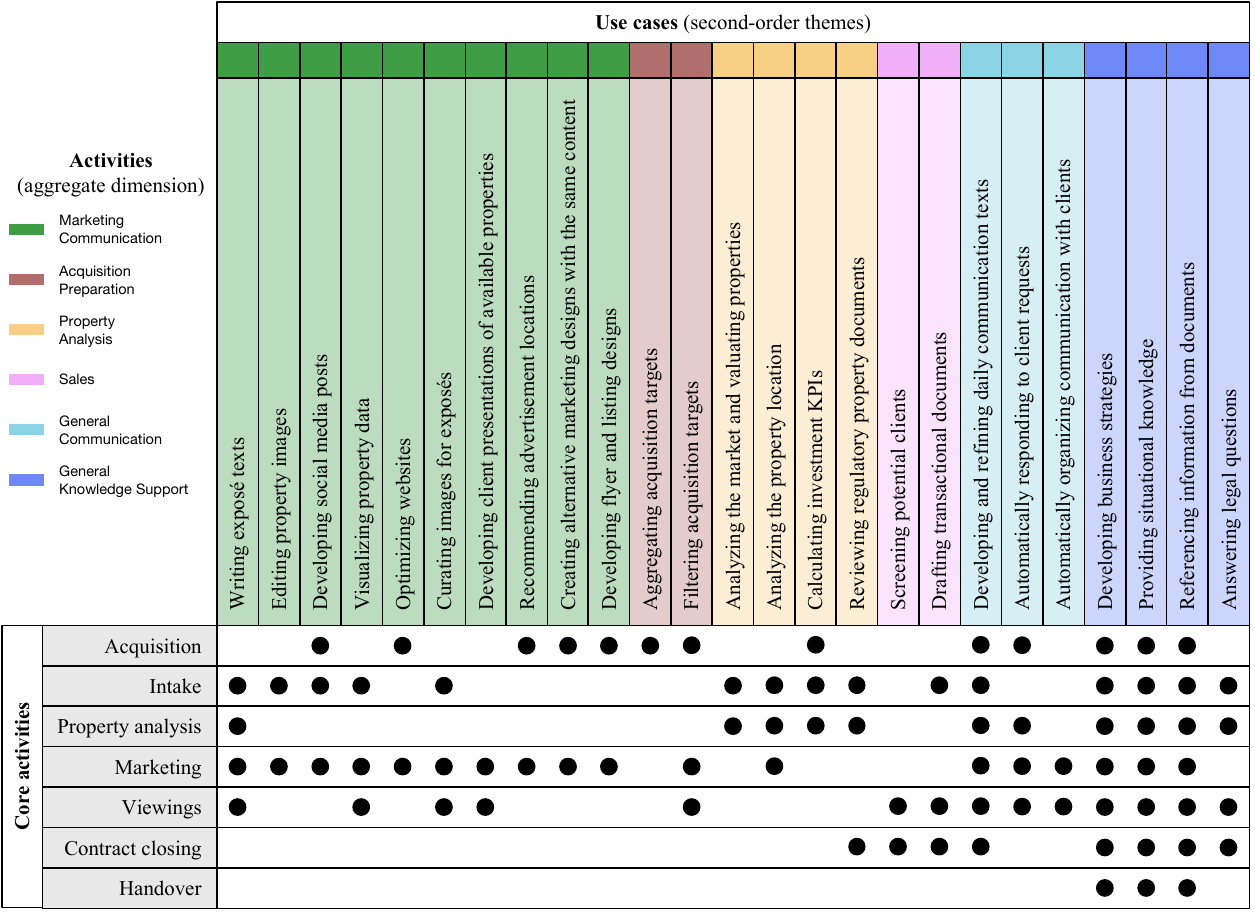}
	\caption{Use cases and activities identified in this work are mapped to the core activities of REAs as defined by~\citet{helfrich_erfolgsstrategien_2021}.}
	\label{fig:mapping}
\end{figure*}

The use cases and corresponding activities that emerged from the interviews can be mapped across the core activities identified by \citet{helfrich_erfolgsstrategien_2021}, as shown in \Cref{fig:mapping}.

\begin{figure*}[tbh]
    \centering
	\includegraphics[width=\linewidth]{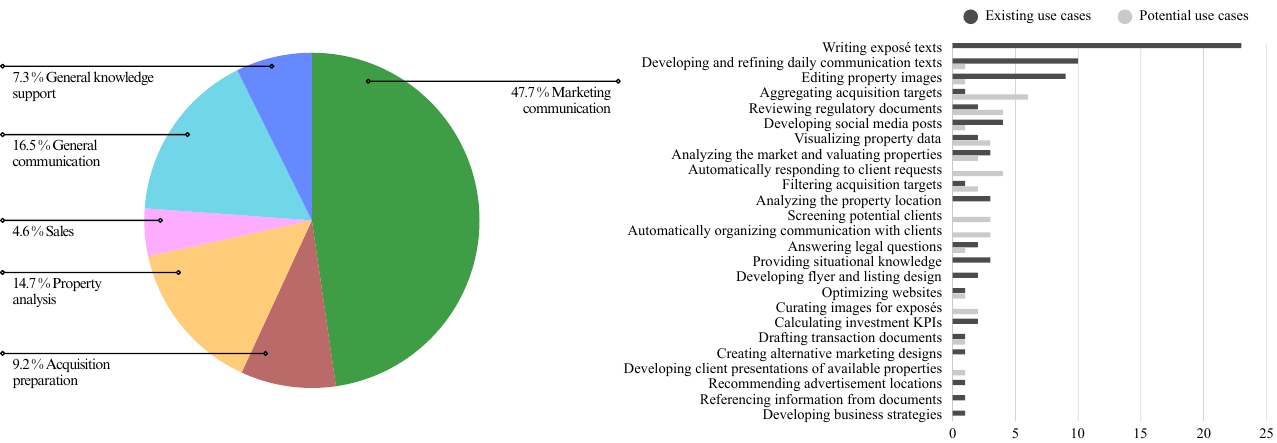}
	\caption{Distribution of participant-level first-order use-case references across aggregate dimensions (left) and distribution of current and potential references across second-order concepts (right).}
	\label{fig:distribution}
\end{figure*}

Marketing, acquisition, and property analysis correspond directly to core REA activities. Sales use cases map to client and contract-related activities, while general knowledge support spans the transaction flow. Based on participant-level first-order use-case references, most use cases relate to marketing communication (47.7 \%) (see \Cref{fig:distribution}), followed by general communication (16.5 \%), property analysis (14.7 \%), acquisition preparation (9.2 \%), general knowledge support (7.3 \%), and sales (4.6 \%). These percentages are descriptive only and should not be read as representative adoption rates. Across the overall set and the subset of current use cases, REAs most frequently refer to writing expos\'e texts, developing and refining daily communication texts, and editing property images. Among potential use cases, aggregation of acquisition targets shows the largest discrepancy between potential and current use and was mentioned by five REAs, underscoring its perceived relevance.

\section{Interpretation and discussion}

Use is reported across all activities, with marketing communication dominating, followed by general communication. We interpret this pattern as reflecting the comparatively low threshold of text-centric tasks: participants described GenAI mainly as a drafting, structuring, and refinement aid in these areas, where outputs can be reviewed quickly by REAs. The two categories differ in work function: marketing communication creates property-facing promotional material, whereas general communication coordinates inquiries, viewings, negotiations, and follow-up. This confirms the U.S. finding that marketing material creation is a central entry point for GenAI adoption~\citep{kriegbaum_chatbots_2024}. However, our data show that the German field is not limited to content generation. Property analysis, acquisition preparation, and document-related knowledge support are also salient because German REM depends on reliable integration of property facts, regulatory documents, and locally distributed market information. Beyond communication, reported use cases include structured information gathering, document interpretation, strategic advice, and ad hoc retrieval. As an author interpretation, lower adoption in these areas appears to reflect heterogeneous data sources, regulatory knowledge requirements, and less standardized outputs. Sales shows the fewest use cases. Here, participants explicitly raised uncertainty about personal data and legal consequences, pointing to compliance concerns under the GDPR and the EU AI Act. Adoption is uneven overall: only expos\'e writing is broadly established, while aggregation of acquisition targets stands out as a high-potential but not yet routine use case. Compared with U.S. evidence, this surfaces cross-portal and document-integration problems as central barriers alongside accuracy, ethics, and human intervention~\citep{kriegbaum_chatbots_2024}. The findings suggest that GenAI fits best where tasks are standardized, audience-facing, and largely textual. Where tasks require verifiable facts, personal data, or binding commitments, participants more often describe GenAI as a drafting or retrieval aid than as an autonomous actor. Some use cases are therefore GenAI-enabled workflows rather than tasks solved by GenAI alone, because complementary technologies may be needed for data access, automation, calculation, or system integration. This matters for visually edited marketing material and numerical KPI calculations, where misleading representations or incorrect outputs can damage client trust and professional reputation. In most reported practices, GenAI drafts, structures, and retrieves while REAs curate, verify, and decide. We therefore describe a partial shift toward editorial judgment and oversight, not a full transformation of the REA role.

\subsection{Practical and theoretical implications}

For practical use, the findings indicate where REAs perceive GenAI as useful and where they remain cautious. Marketing communication appears suitable for low-risk experimentation; acquisition preparation, especially cross-portal aggregation and filtering, can be explored with human review and documented prompts. In domains involving personal data, REAs should use only auditable, compliance-aware processes. Broader implications are tentative because the study did not directly investigate system design, law, or training. Still, reported constraints point to demand for cross-portal aggregation, document retrieval, reviewable outputs, and sector-specific AI literacy around prompt use, output verification, data protection, disclosure of edited content, and when specialized analytics tools are preferable to general-purpose GenAI. The study advances the literature by extending U.S. evidence to German REM and providing a more granular structure of use cases~\citep{kriegbaum_chatbots_2024}. More generally, the findings suggest that GenAI adoption in professional services depends on task standardization, data accessibility, output verifiability, and compliance requirements.

\subsection{Limitations and future work}

The qualitative design and small sample limit statistical generalization. Because the sample includes only REAs with AI experience, the study is suited to mapping practiced and anticipated use cases but cannot explain non-use, resistance, or adoption barriers among REAs who have not yet experimented with AI. Future studies can test and extend these findings with larger and more diverse samples from different regions, including non-users, and with quantitative designs. Risk and legality were not the focus of this work. Several identified applications may raise discrimination or data protection concerns in deployment, especially in screening scenarios. Future research should assess the legal feasibility of specific use cases in light of the EU AI Act and the GDPR. Finally, the study maps what REAs do and what they want to do rather than how to implement these systems. Technical work can evaluate pipelines for the most salient opportunities, including acquisition target aggregation and document understanding in compliance critical contexts. Work on adoption drivers can explain the observed heterogeneity across firms and roles and can support the transition from isolated use to coherent practice.
\section{Conclusion}

This study offers the first empirical account of how REAs in Germany use GenAI in REM. Drawing on semi-structured interviews and an inductive analysis, we find that GenAI already supports activities across the REA workflow. The findings provide an overview of current and potential use cases in German REM and suggest that, in many reported practices, GenAI shifts parts of the work toward review, verification, and oversight. They also identify practical adoption challenges around data access, workflow integration, output verification, and compliance-sensitive tasks. While participants already report GenAI use in REM, the field is at an early stage. Future work should track adoption over time, evaluate governance approaches and AI-literacy programs, and develop implementable pipelines for the high-value opportunities identified here.

\clearpage

\bibliographystyle{ACM-Reference-Format}
\bibliography{references}  %%% Uncomment this line and comment out the ``thebibliography'' section below to use the external .bib file (using bibtex) .

%%% Uncomment this section and comment out the \bibliography{references} line above to use inline references.
% \begin{thebibliography}{1}

% 	\bibitem{kour2014real}
% 	George Kour and Raid Saabne.
% 	\newblock Real-time segmentation of on-line handwritten arabic script.
% 	\newblock In {\em Frontiers in Handwriting Recognition (ICFHR), 2014 14th
% 			International Conference on}, pages 417--422. IEEE, 2014.

% 	\bibitem{kour2014fast}
% 	George Kour and Raid Saabne.
% 	\newblock Fast classification of handwritten on-line arabic characters.
% 	\newblock In {\em Soft Computing and Pattern Recognition (SoCPaR), 2014 6th
% 			International Conference of}, pages 312--318. IEEE, 2014.

% 	\bibitem{keshet2016prediction}
% 	Keshet, Renato, Alina Maor, and George Kour.
% 	\newblock Prediction-Based, Prioritized Market-Share Insight Extraction.
% 	\newblock In {\em Advanced Data Mining and Applications (ADMA), 2016 12th International 
%                       Conference of}, pages 81--94,2016.

% \end{thebibliography}

\end{document}